\documentclass[10pt,journal,compsoc]{IEEEtran}

\usepackage{amsmath,amsfonts,bm}

\def\eqref#1{equation~\ref{#1}}
\def\1{\bm{1}}

\def\rvp{{\mathbf{p}}}
\def\rvq{{\mathbf{q}}}

\def\rvw{{\mathbf{w}}}

\def\rmW{{\mathbf{W}}}
\def\rmX{{\mathbf{X}}}
\def\rmY{{\mathbf{Y}}}

\DeclareMathAlphabet{\mathsfit}{\encodingdefault}{\sfdefault}{m}{sl}
\SetMathAlphabet{\mathsfit}{bold}{\encodingdefault}{\sfdefault}{bx}{n}

\usepackage{times}
\usepackage{epsfig}
\usepackage{graphicx}
\graphicspath{{fig/}}
\usepackage{amsmath}
\usepackage{amsthm}
\usepackage{amssymb}
\usepackage{dsfont}
\usepackage{algorithmic}
\usepackage{color}
\usepackage{soul}
\usepackage{caption}
\usepackage{calc}
\usepackage{enumitem}
\usepackage{calc}
\usepackage{bbold}
\usepackage{algorithm,algorithmic}
\usepackage{siunitx}

\usepackage{tabularx}
\usepackage{booktabs}
\usepackage{multirow}
\usepackage{bbm}

\usepackage{overpic}
\usepackage{tikz}
\usetikzlibrary{calc}
\usetikzlibrary{fit}

\usepackage[pagebackref=false,breaklinks=true,letterpaper=true,colorlinks,bookmarks=false]{hyperref}
\graphicspath{{./img/}}

\newif\ifdraft
\draftfalse
\ifdraft
 \newcommand{\PF}[1]{{\color{red}{\bf PF: #1}}}
 
 \newcommand{\AG}[1]{{\color{blue}{\bf AG: #1}}}
 
 \newcommand{\MK}[1]{{\color{blue}{\bf MK: #1}}}

 \newcommand{\changeSecond}[1]{{\color{blue}#1}}
\else
 \newcommand{\PF}[1]{}
 
 \newcommand{\AG}[1]{}
 
 \newcommand{\MK}[1]{}

 \newcommand{\changeSecond}[1]{{#1}}
\fi

\newcommand{\change}[1]{{#1}}

\newcommand{\RTracer}{{\it RoadTracer}}
\newcommand{\Segm}{{\it Segmentation}}
\newcommand{\SegPath}{{\it Seg-Path}}
\newcommand{\DRoad}{{\it DeepRoad}}
\newcommand{\RCNN}{{\it RCNNU-Net}}

\newcommand{\PolyM}{{\it PolyMapper}}
\newcommand{\CE}{{\it UNet-CE}}
\newcommand{\MSE}{{\it UNet-MSE}}
\newcommand{\HomoT}{{\it Homo-Pre}}
\newcommand{\HomoR}{{\it Homo-Reg}}
\newcommand{\DMT}{{\it DMT}}
\newcommand{\Malis}{{\it ConnLoss}}

\newcommand{\Ours}{{\it Homo-Ours}}
 
\newcommand{\RTD}{{\it RTracer}}
\newcommand{\MAS}{{\it Massachusetts}}
\newcommand{\NEU}{{\it Neurons}}
\newcommand{\BRN}{{\it Brain}}

\newcommand{\APLS}{{\it APLS}}
\newcommand{\TLTS}{{\it TLTS}}
\newcommand{\Betti}{{\it Betti}}
\newcommand{\Junc}{{\it JCT}}

\newcommand{\CCQ}{{\it CCQ}}

\begin{document}
	
	\author{Doruk~Oner,
                Ad\'{e}lie Garin,
				Mateusz~Kozi\'{n}ski,
				Kathryn Hess,
				Pascal~Fua, \textit{Fellow,~IEEE}
	
		\IEEEcompsocitemizethanks{
			\IEEEcompsocthanksitem Doruk Oner and Pascal Fua are with the Computer Vision Laboratory, and Ad\'{e}lie Garin and Kathryn Hess with the Laboratory for Topology and Neuroscience, at \'{E}cole Polytechnique F\'{e}d\'{e}rale de Lausanne, 1015 Lausanne, Switzerland. \protect\\
			E-mail: \{doruk.oner,adelie.garin,kathryn.hess,pascal.fua\}@epfl.ch
			\IEEEcompsocthanksitem Mateusz~Kozi\'{n}ski is with the Institute of Computer Graphics and Vision, TU Graz, 8010 Graz, Austria. 
			E-mail: mateusz.kozinski@icg.tugraz.at
		}%
		\thanks{(Corresponding author: Doruk Oner)}}
	
	\title{Persistent Homology with Improved Locality Information for more Effective Delineation} 
	
	\IEEEtitleabstractindextext{
		
		% !TEX root = ../top.tex
% !TEX spellcheck = en-US

\begin{abstract}

%Persistent Homologies have been successfully used to increase the performance of deep networks trained to detect curvilinear structures and to improve the topological quality of the results. However, existing methods are very global and ignore the location of topological features. In this paper, we introduce an approach that relies on a new filtration function to account for location during network training. We demonstrate experimentally on 2D images of roads and 3D image stacks of neuronal processes that networks trained in this manner are better at recovering the topology of the curvilinear structures they extract.

%\ag{We present a new, more effective way to use Persistent Homology (PH), a method to compare the topology of two data sets, for training deep networks to delineate road networks in aerial images and neuronal processes in microscopy scans. Its essence is in a novel filtration function, derived from a fusion of two existing techniques: thresholding-based filtration, previously used to train deep networks to segment medical images, and filtration with height functions, used before for comparison of 2D and 3D shapes. We experimentally demonstrate that deep networks trained with our Persistent-Homology-based loss yield reconstructions of road networks and neuronal processes that preserve the connectivity of the originals better than existing topological and non-topological loss functions.}

Persistent Homology (PH) has been successfully used to train networks to detect curvilinear structures and to improve the topological quality of their results. However, existing methods are very global and ignore the location of topological features. In this paper, we remedy this by introducing a new filtration function that fuses two earlier approaches: thresholding-based filtration, previously used to train deep networks to segment medical images, and filtration with height functions, typically used to compare 2D and 3D shapes. We experimentally demonstrate that deep networks trained using our PH-based loss function yield reconstructions of road networks and neuronal processes that reflect ground-truth connectivity better than networks trained with existing loss functions based on PH. Code is available at \href{https://github.com/doruk-oner/PH-TopoLoss}{https://github.com/doruk-oner/PH-TopoLoss}.

\end{abstract}

	}
	
	\maketitle

%\input{tex/abstract}
% !TEX root = ../top.tex
% !TEX spellcheck = en-US

%\vspace{-2mm}
\section{Introduction}
%\vspace{-2mm}

In many image segmentation tasks, the topology of the resulting mask is as important as, if not more than, its pixel-wise accuracy. For example, a model of an aortic valve that does not form a ring is biologically implausible. Similarly, networks of curvilinear structures----be they roads in aerial images, blood vessels in Computer Tomography (CT) scans, or dendrites and axons in Light Microscopy (LM) image stacks---should not feature breaks that disrupt connectivity or false connections between disjoint structures. Unfortunately, deep networks trained by minimizing pixel-wise loss functions, such as the cross-entropy or the mean square error, are subject to such mistakes. This is in part because it often takes very few mislabeled pixels to alter the topology significantly with little impact on the pixel-wise accuracy. In other words, it is possible for a network trained in this manner to deliver both a good pixel classification accuracy and an incorrect topology.

Specialized solutions to this problem have been proposed in the form of loss functions that compare the topology of the prediction to that of the annotation. They are effective for specific applications but do not naturally generalize. For example, the perceptual loss of~\cite{Mosinska18} penalizes topological differences between the prediction and the ground truth, but cannot be guaranteed to detect them all. Similarly, minimizing the MALIS loss for segmenting electron microscopy scans~\cite{Turaga09,Funke18} yields better region boundaries but does not penalize interruptions in loopy linear structures.  This has been addressed by~\cite{Oner21a} for delineation of 2D road networks but the proposed solution is not applicable to 3D image stacks. 

Persistent Homology {(PH)~\cite{Edelsbrunner08}}, an elegant approach to describing and comparing topological structure of data, offers the promise to address the connectivity problem in a generic way, both for 2D and 3D images. {Homology is the study of topological features in an object, such as its connected components ($0$-homology classes), loops ($1$-homology classes), and closed surfaces ($2$-homology classes).}
{Persistent homology detects homology classes in objects \emph{filtered} at different \emph{scales}.}
{A homology class that appears at a particular scale and disappears at a larger one is represented by a scale interval called the \emph{persistence interval}. The set of persistence intervals for all the homology classes characterizes the overall topology of the structure.} 
 It can be represented by a  {\it persistence diagram}. The similarity of these diagrams across two different structures can then be used to quantify their topological similarity. This has been successfully exploited to train deep networks for delineation~\cite{Hu19b}, image segmentation{~\cite{Hu19b,Clough19,Clough20}} and crowd counting~\cite{Abousamra21}. 

\begin{figure*}[ht]
	\centering
	\begin{tabular}{cccc}
	\hspace{-3mm}\includegraphics[height=2.8cm]{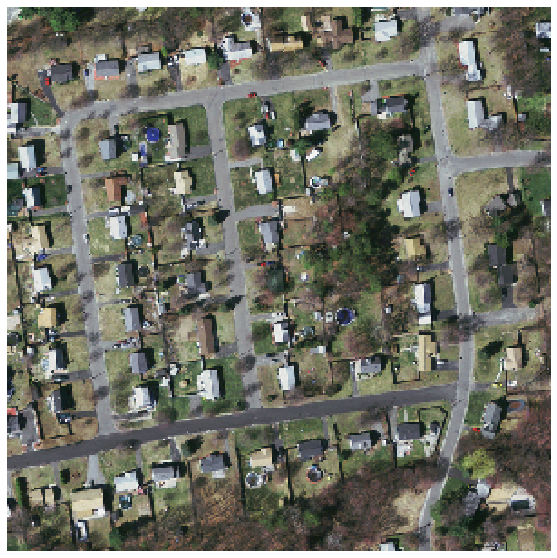} &
	\hspace{-3mm}\includegraphics[height=2.8cm]{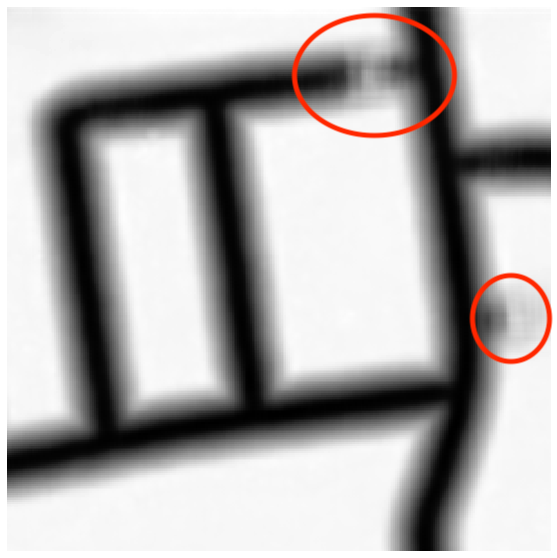}&
	\hspace{-3mm}\includegraphics[height=2.8cm]{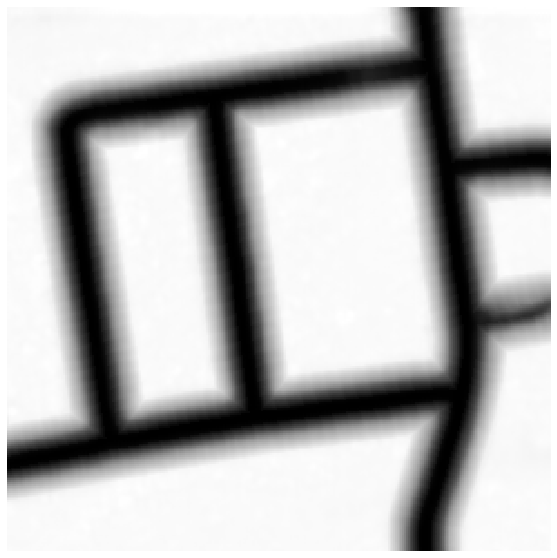}&
	\hspace{-3mm}\includegraphics[height=2.8cm]{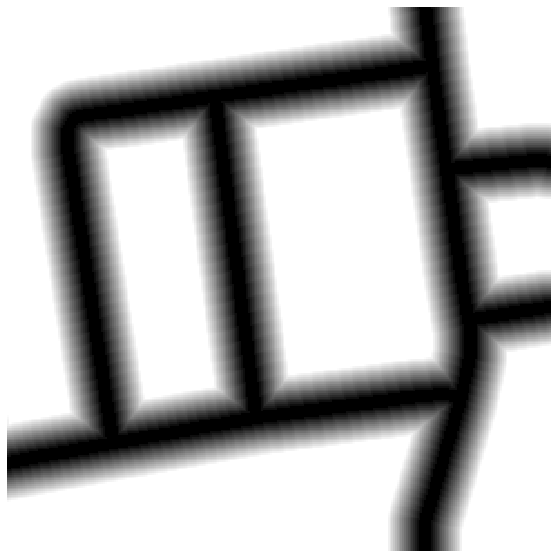}\\
	
	\hspace{-3mm}\includegraphics[height=2.8cm]{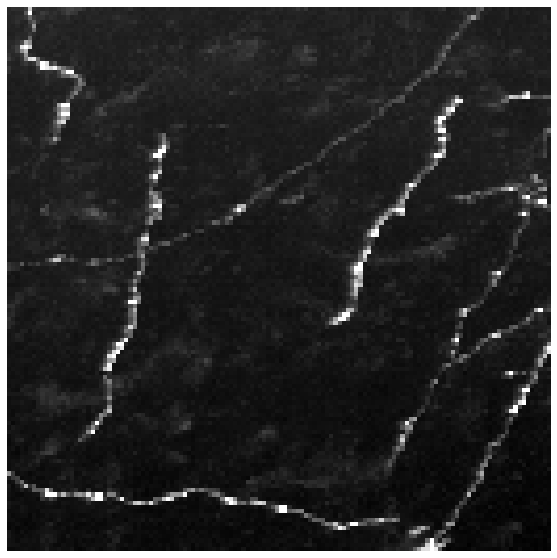} &
	\hspace{-3mm}\includegraphics[height=2.8cm]{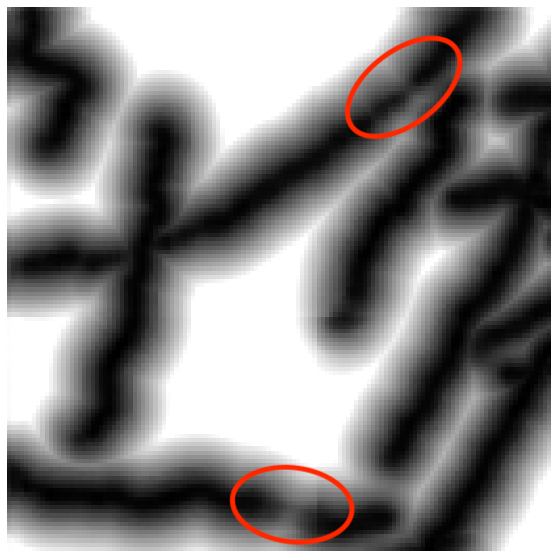}&
	\hspace{-3mm}\includegraphics[height=2.8cm]{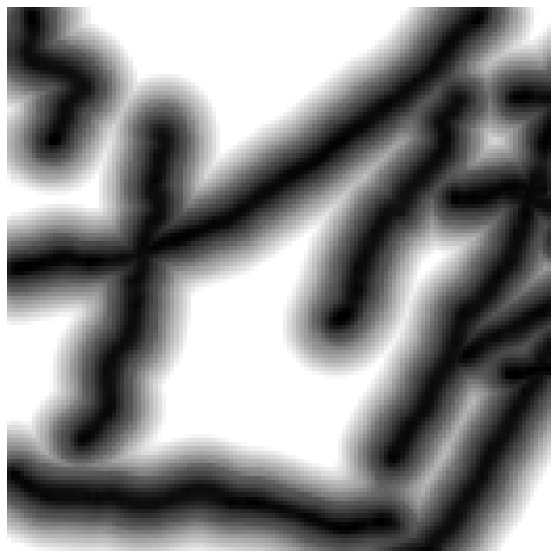}&
	\hspace{-3mm}\includegraphics[height=2.8cm]{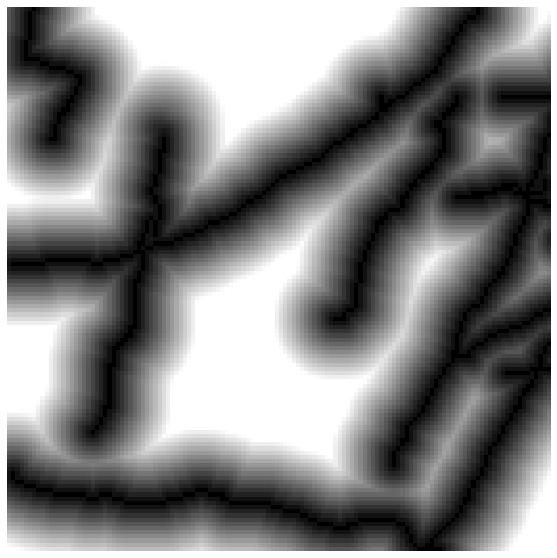}\\
	(a)&(b)&(c)&(d)
	\end{tabular}
	\vspace{-3mm}
\caption{\small {\bf 2D and 3D delineation. } (a) Aerial image and slice of a microscopy stack. (b) A network trained using a standard homology-based loss yields road and neurite interruptions. (c) One trained using our localized loss is more topologically accurate and produces predictions that closely resemble the ground truth (d).}
       \vspace{-3mm}
\label{fig:teaser}
\end{figure*}

We show that these methods fail to unleash the full power of persistent homology,
because they discard too much information about the structure of the prediction and the annotation when encoding them in the form of persistence diagrams. 
As shown in Fig.~\ref{fig:teaser}, this can result in networks that still fail to enforce the proper topology.
To remedy this, we introduce a new approach to computing persistence diagrams that increases their descriptive power, as shown in Fig.~\ref{fig:diagrams}. Our main contribution is a novel filtration technique {that combines two approaches to filtration commonly used in topological data analysis (TDA): thresholding-based-filtration~\cite{Clough19,Clough20,Hu19b} and the height function~\cite{Turner14}.}
It yields a loss function that can be used for both 2D and 3D images and significantly improves performance compared to state-of-the-art topological methods, as we will demonstrate in our experiments.

%However, existing techniques do not unleash the full power of persistent homology because the persistent diagrams are global image {descriptors} that ignore the {location} of the topological features, which reduces their descriptive power. As shown in Fig.~\ref{fig:teaser}, this can result in networks that still fail to enforce the proper topology. This is because, when training a deep network, a persistence-based loss can be low even if the network predicts a structure that is quite different from the ground-truth. To remedy this, we introduce a new approach to computing persistence diagrams that takes location into account and increases their descriptive power, as shown in Fig.~\ref{fig:diagrams}. Our main contribution is a novel filtration technique {that combines two approaches to filtration commonly used in topological data analysis (TDA): Thresholding based filtrations and using height functions.} \PF{Add a couple of references here.}
%{It yields a loss function that can be used for both 2D and 3D images and significantly improves performance compared to state-of-the-art topological methods, as will demonstrate in our experiments.}

% !TEX root = ../top.tex
% !TEX spellcheck = en-US

%\input{fig/diagrams}

%--------
% OLD
%--------

%two existing techniques: thresholding-based filtration, previously used to train deep networks to segment medical images, and filtration with height functions, used before for comparison of 2D and 3D shapes. We experimentally demonstrate that deep networks trained with our Persistent-Homology-based loss yield reconstructions of road networks and neuronal processes that preserve the connectivity of the originals better than existing topological and non-topological loss functions.

% !TEX root = ../top.tex
% !TEX spellcheck = en-US

%\vspace{-2mm}
\section{Related work}
%\vspace{-2mm}

Training a deep network that produces topologically correct segmentations has typically been done by designing loss functions that, when minimized, favor plausible topology. In this section, we briefly review first those that do not rely on Persistent Homology, and then those that do. 

\subsection{Losses Designed to Enforce Topological Correctness.}

Several such losses have been proposed already to go beyond pixel-wise classification accuracy by encoding more global properties. In~\cite{Li20c}, the connectivity between neighboring pixel pairs is used as an additional source of supervision. This approach has been shown to improve connectivity, but since disconnections or false connections are not penalized explicitly, there is no guarantee it captures all such errors. The perceptual loss of~\cite{Mosinska18} is based on the assumption that a pre-trained neural network can capture differences of connectivity between the prediction and the ground-truth. However, even though it has been shown experimentally to improve the topology of masks produced by a deep net, there is no guarantee that this assumption holds in general. Making the Rand index of segmentations produced by the network similar to that of ground truth ones~\cite{Turaga09,Funke18} helps when modeling tree-like structures, both in 2D and in 3D, but cannot prevent disconnections in loopy structures. This shortcoming has been addressed by~\cite{Oner21a} by detecting disconnections of 2D loopy structures as interconnections of background regions, but the proposed solution does not generalize to 3D. 

\subsection{Losses that rely on Persistent Homology}

Persistent Homology (PH)~\cite{Edelsbrunner00,Zomorodian04} is a well-established topological data descriptor.
{
One of its important applications is comparing the topological structures of binary images, for example by enforcing the correct Betti number on binary masks resulting from inference in Markov Random Fields~\cite{Chen11a}. More recently, it has been shown that persistence diagrams can also be computed for grayscale images and differentiated with respect to the pixel values~\cite{Hu19b,Clough19,Gabrielsson20,Leygonie21,Carriere21}. Hence, they can be incorporated into loss terms and used to train deep networks. In this vein,  a loss term that enforces a sequence of desired Betti numbers on the predicted segmentation was introduced in~\cite{Clough19}. This approach was further extended to a loss function that tends to equalize the Betti number of the prediction and the ground truth~\cite{Clough20}. In a similar vein, the loss term of~\cite{Hu19b} relies on comparing  persistence diagrams of the prediction and the ground truth, where the persistence diagrams are obtained by thresholding. As discussed in the next section, for binary ground truth this results in degenerate persistence diagrams that only encode the Betti number. Thus,  this approach  can be interpreted as equalizing the Betti numbers of the prediction and the ground truth, as in~\cite{Clough20}. This was improved upon in~\cite{Wang20e} by applying PH to distance maps instead of binary annotations or class affinity maps. We show in the next section that this makes the loss function better at detecting and penalizing topological errors. Unfortunately, even this improved technique is susceptible to incorrectly matching the persistence diagrams of the prediction and the ground truth because it throws away location information. By incorporating such information into our diagrams, our method makes them more informative and alleviates this problem. 
}

{It has also been proposed to detect disconnections in predicted 2D and 3D structures using Discrete Morse Theory~\cite{Hu21b}. Topological features that are inconsistent with the ground truth are then penalized in the loss function.  However, when the annotations lack spatial precision, which is often the case for neurite and road centerline annotation like the ones studied here, ground-truth inaccuracies may confuse the network. By contrast, our technique allows for considerable misalignment between the prediction and the ground truth.}

% !TEX root = ../top.tex
% !TEX spellcheck = en-US

%\vspace{-2mm}
\section{Method}
%\vspace{-2mm}
\label{sec:method}

We first introduce Persistent Homology and its application to characterizing two-dimensional images and three-dimensional image stacks. As PH provides global descriptors that ignore the location of {topological features}, we then introduce our approach to accounting for it. 

%\vspace{-2mm}
\subsection{Persistent Homology}
%\vspace{-2mm}

% !TEX root = ../top.tex
% !TEX spellcheck = en-US

\begin{figure*}[htb!]
\center
\includegraphics[width=0.9\textwidth]{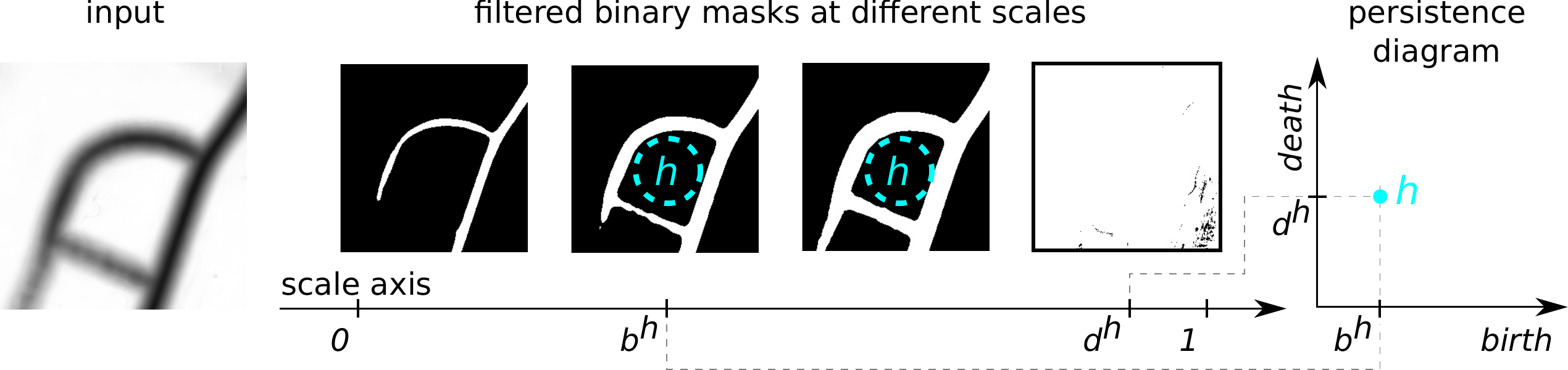}
\vspace{-2mm}
\caption{
\small {\bf Filtration.} 
When the distance map shown on the left is filtered by thresholding, the loop $h$ emerges at scale $b^h$ and is filled at scale $d^h$. This gives rise to the point $(b^h,d^h)$ in the persistence diagram shown on the right. Here, thresholding means retaining all pixels whose value is lower than the threshold.
}
\vspace{-4mm}
\label{fig:filtration}
\end{figure*}

In the interest of simplicity, we introduce PH for binary images and image stacks, where {homology classes} are limited to connected components, loops, and closed surfaces. We refer the interested reader to the review{~\cite{Edelsbrunner00}} for a more general treatment, applicable to non-image and higher-dimensional data.

{At the heart of PH is detecting homology classes---connected components, loops, closed surfaces---at many different scales. The ones that exist over a wide scale range are called persistent and deemed more likely to represent true features, as opposed to sampling artifacts or noise.
%The lifespan of the homology classes throughout this process is summarized in what is called a persistence diagram.
Here, scale has a very specific meaning. It refers to the parameter of a filtration function $F$ that is applied to an image $\rmY$ to produce} {topological objects called \emph{cubical complexes}. They arise when filtering images and their properties are described in~\cite{Garin20,Bleile21} for instance. A reader not familiar with algebraic topology can think of them as binary masks.} 
%In the topology literature, this binary mask is called a cubical complex.
%Their construction of these complexes is described in~\citep{Garin20},
%In~\citep{Garin20} the authors explain the two possibilities to build cubical complexes for images and their consequences on the corresponding persistence diagrams. Note that we use the library Gudhi{~\citep{Maria14}} to compute persistence diagrams from images, therefore the corresponding image adjacency is the indirect one \AG{ cite https://arxiv.org/abs/2005.04597 -- done}.}
The masks obtained for different scales form a sequence {of inclusions, that is,} for a pair of scale parameters $s_1 < s_2$, the mask $F(\rmY,s_1)$ is entirely contained within the mask $F(\rmY,s_2)$. The simplest example of a function for filtering grayscale images is thresholding, where the threshold acts as the scale, as shown in Fig.~\ref{fig:filtration}. 

As the scale changes, {homology classes} in the filtered cubical complex emerge and disappear. To capture this, the scale range is sampled from small to large, the image is filtered at the selected scale values, {homology classes} in the resulting binary masks are detected algebraically~{\cite{Edelsbrunner00}}, and correspondence is established between the {homology classes} found at consecutive scales. For each {class}, this yields a pair $(b,d)$, where $b$ is the scale at which the homology {class} appears and $d$ the scale at which it disappears. We will refer to them as {\it birth} and {\it death} times and to the interval $[b,d]$ as the {\it persistence interval} of the homology {class}. The set $P_{\rmY}=\{(b^h,d^h)\}_{h \in H_{\rmY}}$, where $H_{\rmY}$ is the set of all {homology classes} found in {the filtered} image $\rmY$, is called the {\it persistence diagram} of $\rmY$, and was first introduced by~\cite{Barannikov94}. In practice, we use the Gudhi library~\cite{Maria14} to compute persistence diagrams from images. Fig.~\ref{fig:filtration} depicts the birth and death of a specific homology {class}. 

To compare images $\rmY_1$ and $\rmY_2$, one-to-one matching is performed between their persistence diagrams, $P_{\rmY_1}$ and $P_{\rmY_2}$, with the cost of matching a homology $g\in H_{{\rmY}_1}$ to a homology $h\in H_{{\rmY}_2}$ set to $c_{g,h}=(b^{h}-b^{g})^2+(d^{h}-d^{g})^2$ and {the cost of leaving an interval $[b,d)$ unmatched is set to the distance between the point $(b,d)$ and the diagonal in $\mathbb R^2$}. The optimal matching can be found using the Hungarian algorithm. Its cost that we denote as $C({\rmY_1,\rmY_2})$ quantifies the topological discrepancy between ${\rmY}_1$ and ${\rmY}_2$ by penalizing differences between corresponding {homology classes} and ones that only appear in either ${\rmY}_1$ or ${\rmY}_2$.  

%\vspace{-2mm}
\subsection{Training Deep Networks using PH}
%\vspace{-2mm}

Let  $f$ be a network that associates to an input image $\rmX$ a segmentation mask $\rmY = f(\rmX)$ such that for all pixels or voxels $p \in \rmY$, $0\le \rmY[p]\le 1$ and let $\hat{\rmY}$ be the corresponding ground-truth mask. A natural idea then is to train $f$ by minimizing 
\begin{equation} 
L_{\rm{tot}}(\rmY,\hat{\rmY})=L(\rmY,\hat{\rmY})+\alpha C({\rmY},\hat{\rmY}) \; ,
\label{eq:totloss}
\end{equation}
where $L$ is the standard loss function, either the Mean Square Error, or the Cross Entropy, and $\alpha$ is a hyper-parameter, which we set to 0.01 in practice. This is possible because $C$ is sub-differentiable with respect to its inputs when filtration is achieved by thresholding, as shown in~\cite{Hu19b,Clough19,Leygonie21}. However, when the ground truth $\hat{\rmY}$ is binary, as it often is, all structures emerge at scale zero and disappear at scale one. Hence,  as shown in Fig.~\ref{fig:diagrams}(a) the persistence intervals all are $[0,1]$. In other words, all points in a ground truth persistence diagram are located in its upper-left corner, and the only difference between diagrams obtained for annotations of different images is the number of points they contain. Unlike in classical applications of PH~\cite{Edelsbrunner00}, where persistence diagrams serve as rich topological descriptors, such diagrams only encode the Betti number of the annotation. An approach to handling this difficulty is to replace the binary ground truth by its distance transform that can be thresholded over a wide range of threshold values to create different binary masks~\cite{Wang20e}. Unfortunately, computing the persistence diagram of a ground truth distance transform still yields persistence diagrams in which the topological features of the original, binary ground truth are spread along the `death' axis but not along the `birth' one: The distance value at the structures themselves is zero and, as a result, all the loops of the ground truth mask appear as soon as the scale value becomes positive. As shown in Fig.~\ref{fig:diagrams}(b), this may lead to erroneous matches between persistence diagrams, which encourages the deep network to produce wrong segmentations. 
%Moreover, the location of {homology classes} within the image is ignored. This is sub-optimal, because the predicted {topological features} should not be too far from the ground truth ones.

%\vspace{-2mm}
\subsection{{Accounting for the Location of Topological Features during Filtration}} 
%\vspace{-2mm}
% !TEX root = ../top.tex
% !TEX spellcheck = en-US

\begin{figure*}[htb!]
	\centering
	\setlength{\tabcolsep}{2pt}
	\begin{tabular}{@{}c c c c c c@{}}		
		& input & \multicolumn{3}{c}{\small filtered binary masks} & pers.\ diag.\\
	    	%\rotatebox{90}{Binary Masks} & 
		\raisebox{-0.25\height}{\rotatebox{90}{\makebox[2.25cm][c]{\small prediction} \makebox[2.25cm][c]{\small ground truth}}} &
		\raisebox{-0.25\height}
		{\centering\includegraphics[height=4.3cm]{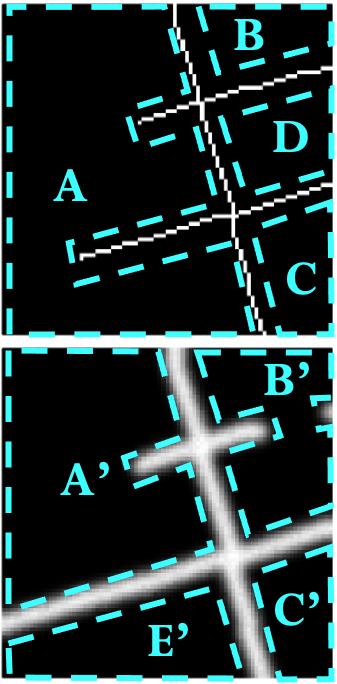}} 
		& \raisebox{-0.25\height}
		{\includegraphics[height=4.3cm]{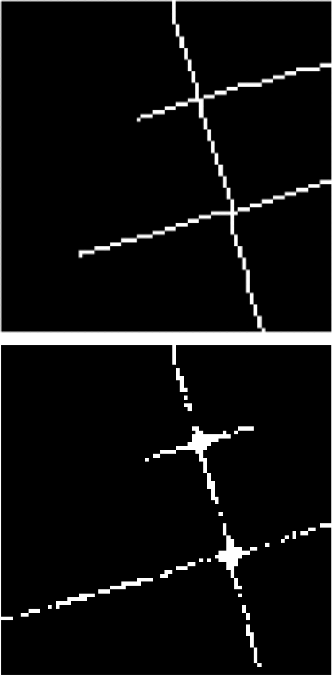}}
		& \raisebox{-0.25\height}
		{\includegraphics[height=4.3cm]{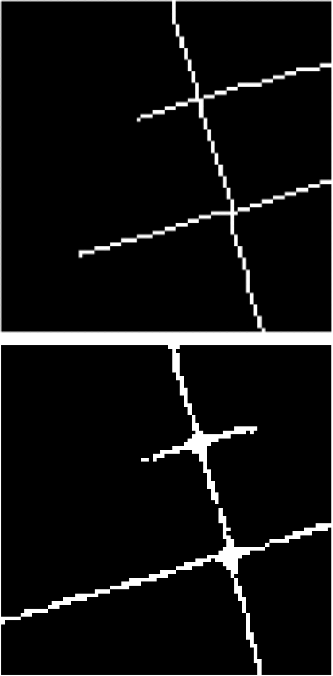}}
		& \raisebox{-0.25\height}
		{\includegraphics[height=4.3cm]{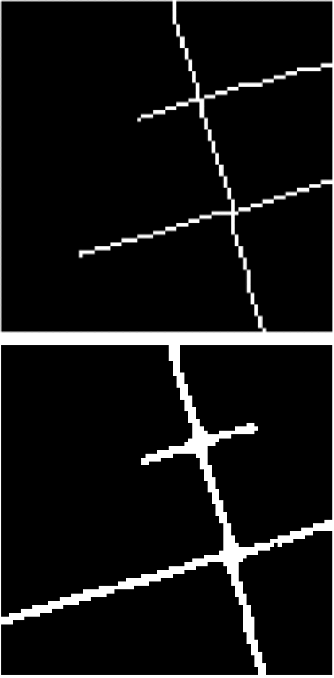}}
		& \raisebox{-0.25\height}
		{\includegraphics[height=4.3cm]{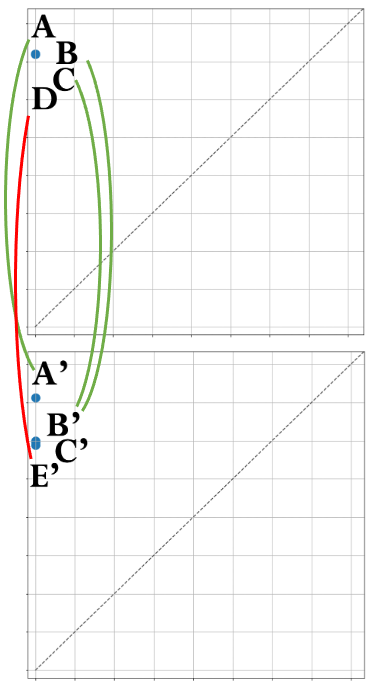}}
		\\
		\multicolumn{6}{>{ \arraybackslash}p{0.9\textwidth}}{\small (a) \textbf{Filtration by thresholding} binary ground truth and predicted class affinity maps. Here, filtration involves decreasing the threshold from 1 to 0, and  retaining the pixels greater than the threshold. Note, that the binary masks resulting from filtering the ground truth at different scales are all the same and that all points in the ground truth persistence diagram ({\it top-right}) coincide. This results in erroneous matches between the predicted and ground truth {homology classes}. Minimizing a loss function based on such a filtration can magnify the errors.} \\
		%&
		%\rotatebox{90}{Distance Maps} & 
		\raisebox{-0.25\height}{\rotatebox{90}{\makebox[2.25cm][c]{\small prediction} \makebox[2.25cm][c]{\small ground truth}}} &
		\raisebox{-0.25\height}
		{\includegraphics[height=4.3cm]{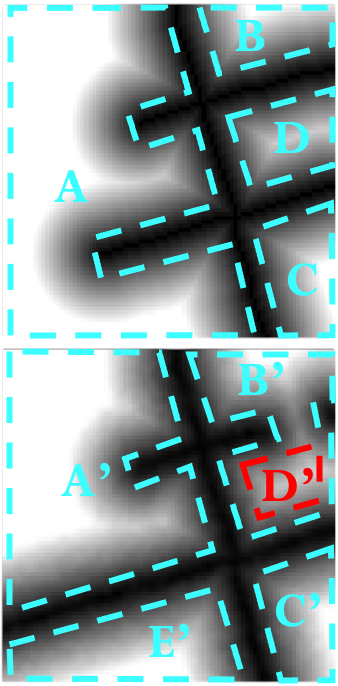}} 
		& \raisebox{-0.25\height}
		{\includegraphics[height=4.3cm]{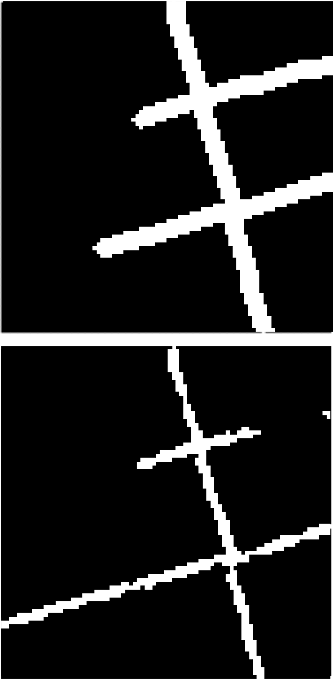}}
		& \raisebox{-0.25\height}
		{\includegraphics[height=4.3cm]{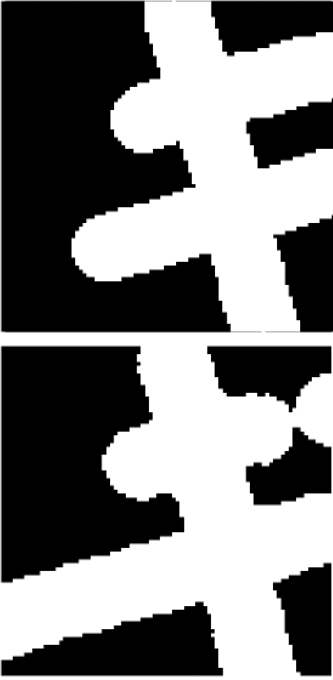}}
		& \raisebox{-0.25\height}
		{\includegraphics[height=4.3cm]{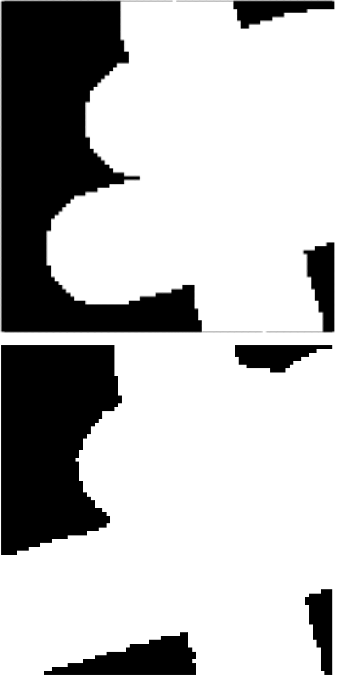}}
		& \raisebox{-0.25\height}
		{\includegraphics[height=4.3cm]{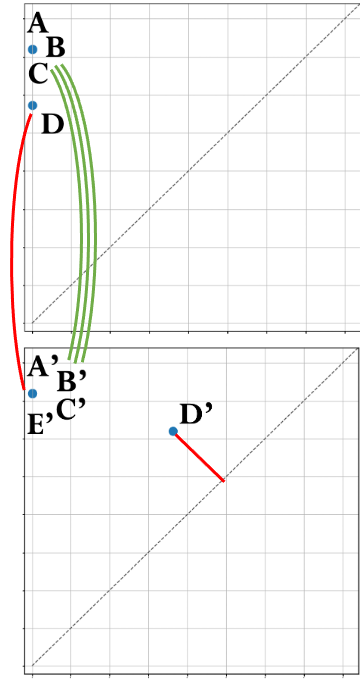}}
		\\
		\multicolumn{6}{>{\arraybackslash}p{0.9\textwidth}}{\small (b) \textbf{Filtration by thresholding distance maps} distributes the topological features of the ground truth along the vertical but not the horizontal axis. This still results in erroneous matching between the predicted and ground truth {homology classes}: Loop D' in the prediction emerges when the threshold is high enough to make the road break disappear. Hence, it remains unmatched and the E' loop created by the false positive road is matched to the ground truth loop D.} \\
		\raisebox{-0.25\height}{\rotatebox{90}{\makebox[2.25cm][c]{\small prediction} \makebox[2.25cm][c]{\small ground truth}}} &
		\raisebox{-0.25\height}
		{\includegraphics[height=4.3cm]{img/comp_im_dist}} 
		& \raisebox{-0.25\height}
		%{\includegraphics[height=4.5cm]{img/comp_th1_ours}}
		{
                  \begin{overpic}[height=4.3cm]{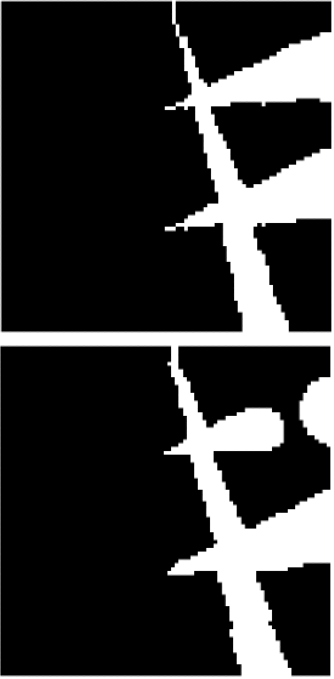}
                    \put(20,72.5){\linethickness{1pt}\color{gray!80}\vector(-3,1){15}}
                    \put( 5,67.5){\tiny\color{gray!80}height axis}
                  \end  {overpic}
                }
		& \raisebox{-0.25\height}
		{\includegraphics[height=4.3cm]{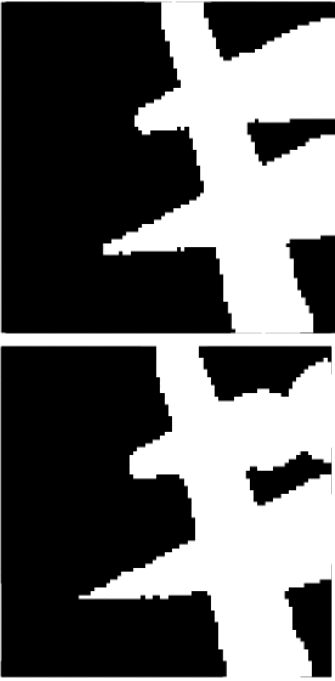}}
		& \raisebox{-0.25\height}
		{\includegraphics[height=4.3cm]{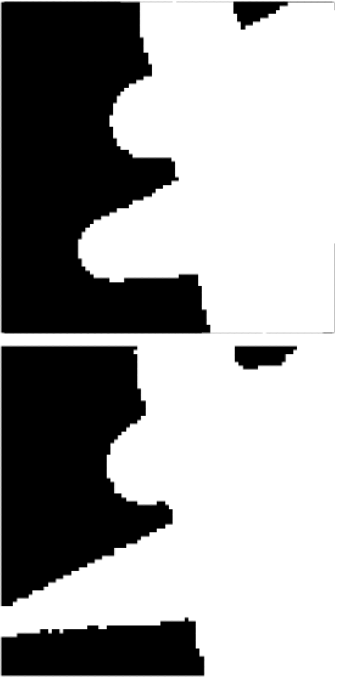}}
		& \raisebox{-0.25\height}
		{\includegraphics[height=4.3cm]{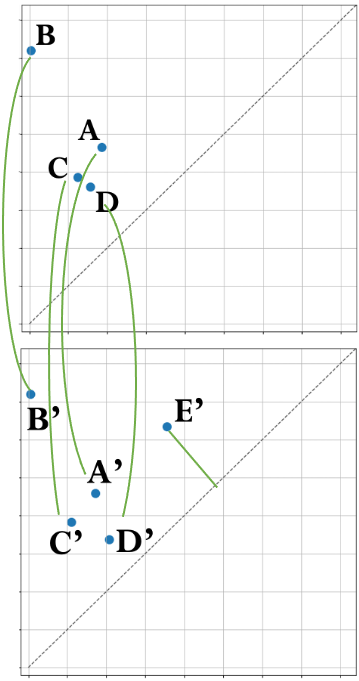}}
		\\
		\multicolumn{6}{>{\arraybackslash}p{0.9\textwidth}}{\small (c) \textbf{Our localized filtration of distance maps} distributes the persistence diagram of the ground truth across the plane, promoting correct matches between predicted and ground truth {homology classes}. \changeSecond{The gray arrow represents the direction of the height axis used by the filtration function.}} 
	\end{tabular}
%	\vspace{-3mm}
	\caption{
		\small {\bf Comparing filtration functions on synthetic data.}
The binary ground truth road annotation ({\it top-left} in each table part) contains four loops, marked with cyan dashed lines. We synthesized a predicted class affinity map ({\it bottom-left} in each part) by extending one road to the left and interrupting another. In consequence, loop B and D from the ground truth are joined into B' in the prediction, and A is split into A' and E'. For each filtration method, we show binary masks resulting from filtration at different scales, pairs of persistence diagrams, and their optimal matches.
}
%\vspace{-5mm}
\label{fig:diagrams}
\end{figure*}

%To remedy the above-mentioned drawbacks of traditional PH, our goal is therefore to spread the persistence diagrams along {\it both} dimensions while also accounting for where in the image the {homology classes} are. 

\changeSecond{
Our goal is therefore to prevent erroneous matches between topological features of the prediction and of the ground truth. To this end, we want to use the features' image location to characterize them. However, re-defining the matching cost to include a position-dependent term would be difficult, because topological features extend across the scale-space, and because there is no natural notion of distance between them. Hence, instead of modifying the matching cost, we propose a new filtration function that distinguishes features at different positions. }
We draw our inspiration from a filtration technique called the height function~\cite{Turner14}. It was originally designed for three-dimensional meshes and can be applied to binary images by assigning to each pixel a {\it height} value that is the coordinate of its projection along a selected straight line. Filtration is carried out by forming binary masks made of pixels whose height is smaller than the scale parameter {\cite{Garin19}}. As the scale is increased, the binary image is revealed in scan-lines perpendicular to the height axis, one scan-line at a time. The birth and death times are the heights of pixels responsible for the emergence and disappearing of {homology classes}. As a result, the persistence diagram contains partial information about the location of {topological features}. Moreover, both birth and death times of different {homology classes} are distributed across scales. 
{
Additionally, it has been shown that a binary image can be reconstructed from as few as four persistence diagrams obtained with height functions with well-chosen directions~\cite{Betthauser18}.
A height function is only defined for binary images, but the abovementioned result inspired us to extend its definition by combining it with thresholding distance maps.
}
Given a scale $s$, the value of the filtered binary mask at coordinates $\rvp$ is taken to be
\begin{equation} \label{eq:filtration}
F(\rmY,s)[\rvp]=\mathbb{1}(\rmY[\rvp]+\rho(\rvp)<s)\;,
\end{equation}
where $\mathbb{1}(\cdot)$ evaluates to one if the condition in the bracket is satisfied and to zero otherwise. In essence, this amounts to thresholding the sum of the height function $\rho$ and the pixel values. 
{From the perspective of TDA, such combination of two filtration functions can be seen as a line in the fibered barcode defined by~\cite{Carriere20}.}

In its simplest form, $\rho$ is a linear function of pixel coordinates, and the region highlighted for any $s$ extends along a line perpendicular to the height axis, as shown in Fig.~\ref{fig:diagrams}(c). But other forms of $\rho$ are also possible. We tested
\begin{itemize}
\item linear functions $\rho(\rvp)=\rvw^\intercal \rvp$, where $\rvw$ is a two-vector hyper-parameter encoding the orientation of the height axis and the slope of the height function; 
\item a scaled distance to a point $\rvq$ in the image,  $\rho(\rvp)=a \|\rvp-\rvq\|_2$, where $\rvq$ and $a$ are hyper-parameters;
\item the square of the height function $\rho(\rvp)=\rvp^\intercal \rmW \rvp$, where $\rmW=\rvw^\intercal\rvw$, and $\rvw$ is the hyper parameter encoding the slope of the function and the orientation of the height axis;
\end{itemize}
The function $\rho$ introduces {localization information of the topological features} into the {persistence diagram}. This is illustrated by Fig.~\ref{fig:diagrams} where different values of the scale parameter make {homology classes} appear in different parts of the image. But, because the scale parameter must be a scalar, it can only {pinpoint location of topological features} in 2D or 3D images along one direction. This could be addressed by evaluating the loss function many times for many different orientations of the height axis, or more generally, for many different hyper-parameters of $\rho$. {This approach is legitimized by the theoretical result of~\cite{Betthauser18} that states that four well chosen filtration directions suffice to completely represent a binary image. The problem of combining a number of different filtration functions is known in topological literature as multipersistence~\cite{Carlsson09}.
But current multipersistence techniques are not easily plugged into a deep learning framework for lack of results on their differentiability.  Moreover, filtering the data along multiple directions would considerably slow down the training.} Instead, we randomly draw the hyper-parameters of the height function at each training iteration. We show in the supplementary material that, in practice, the simple linear function performs best.

% !TEX root = ../top.tex
% !TEX spellcheck = en-US

%\vspace{-2mm}
\section{Experiments}
%\vspace{-2mm}

\change{
We first demonstrate that our loss function correlates with the number of topological errors better than standard PH-based losses. We then evaluate its performance in training deep networks to delineate road networks and neuronal arborizations. }

% !TEX root = ../top.tex
% !TEX spellcheck = en-US

\begin{figure*}[htb!]
	\centering
	\begin{tabular}{@{} c c c c @{}}
		\includegraphics[width=0.18\textwidth]{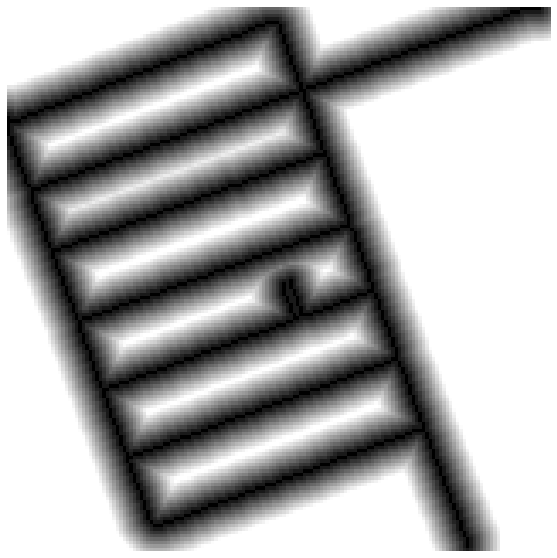} &
		\includegraphics[width=0.18\textwidth]{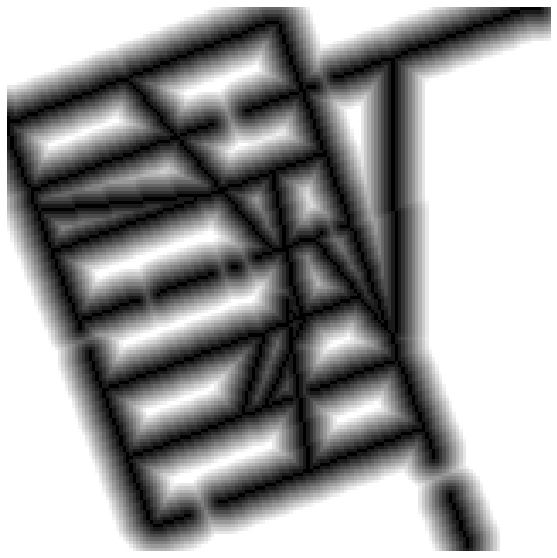} &
		\includegraphics[width=0.18\textwidth]{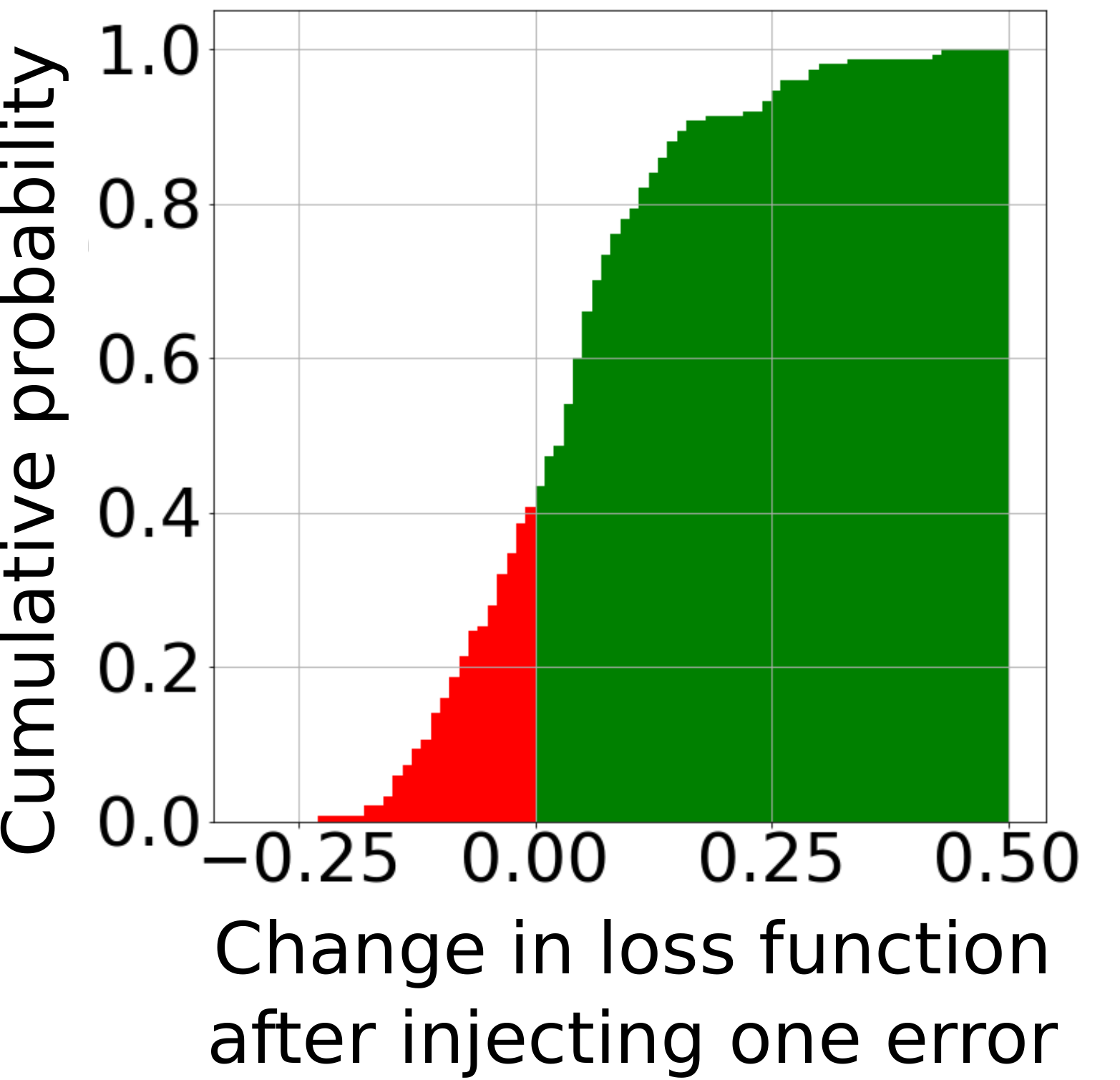} &
		\includegraphics[width=0.18\textwidth]{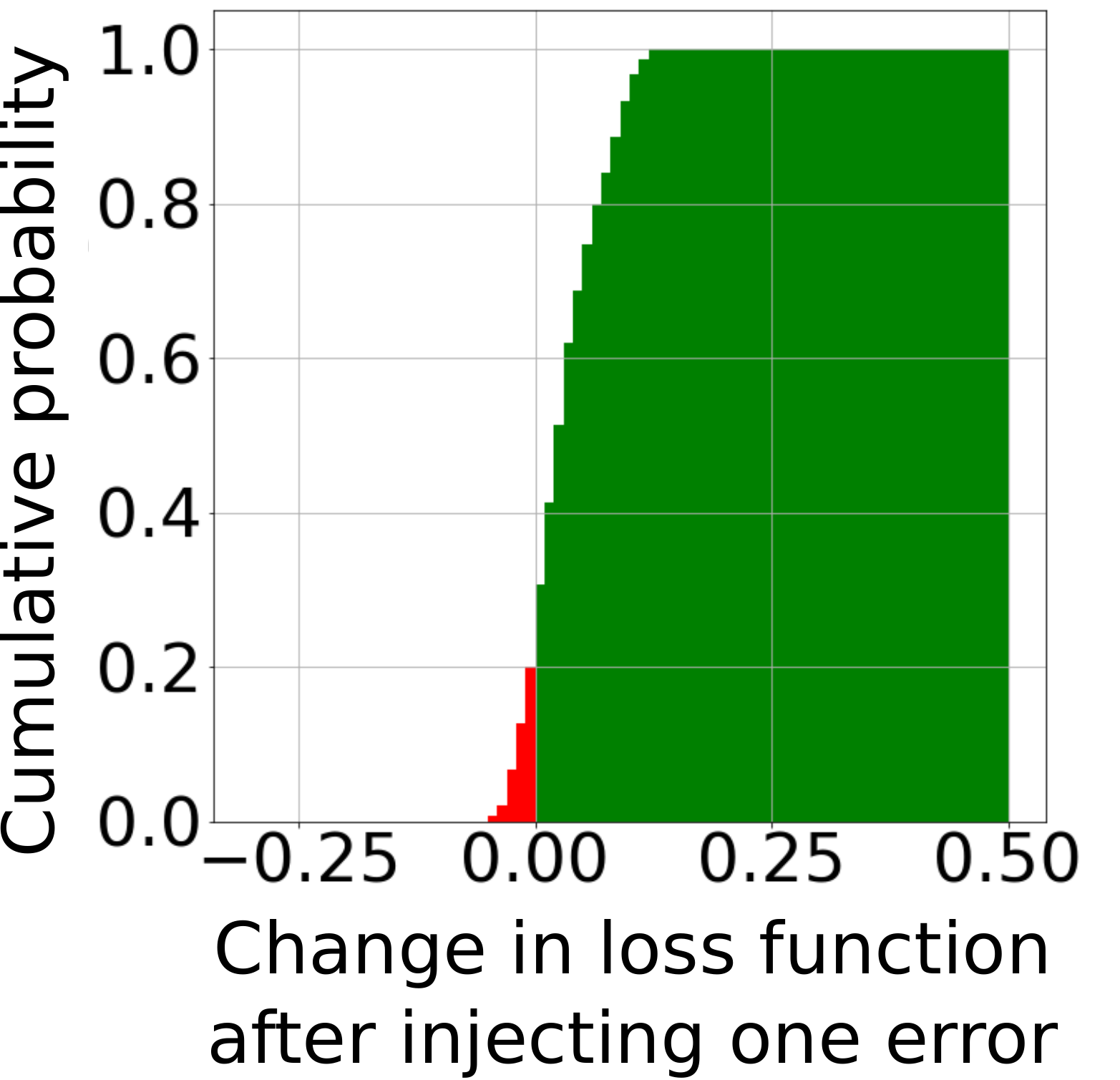} 
		\\
		\includegraphics[width=0.18\textwidth]{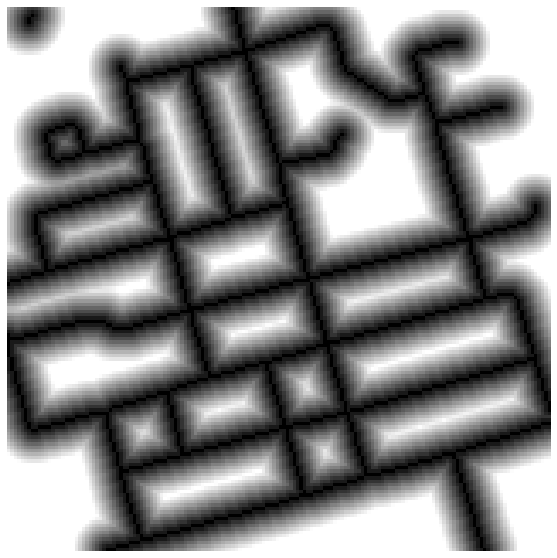} &
		\includegraphics[width=0.18\textwidth]{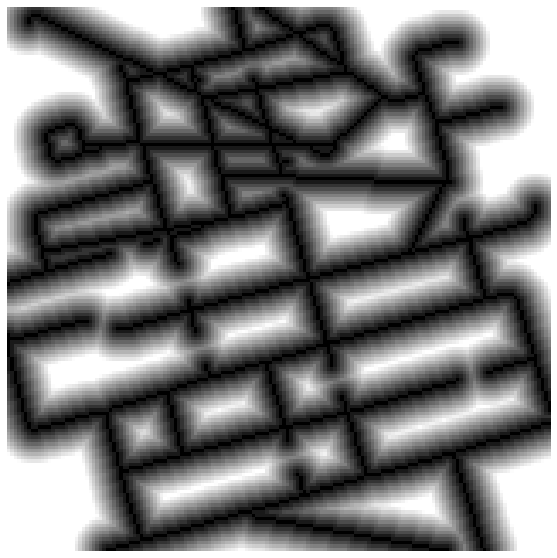} &
		\includegraphics[width=0.18\textwidth]{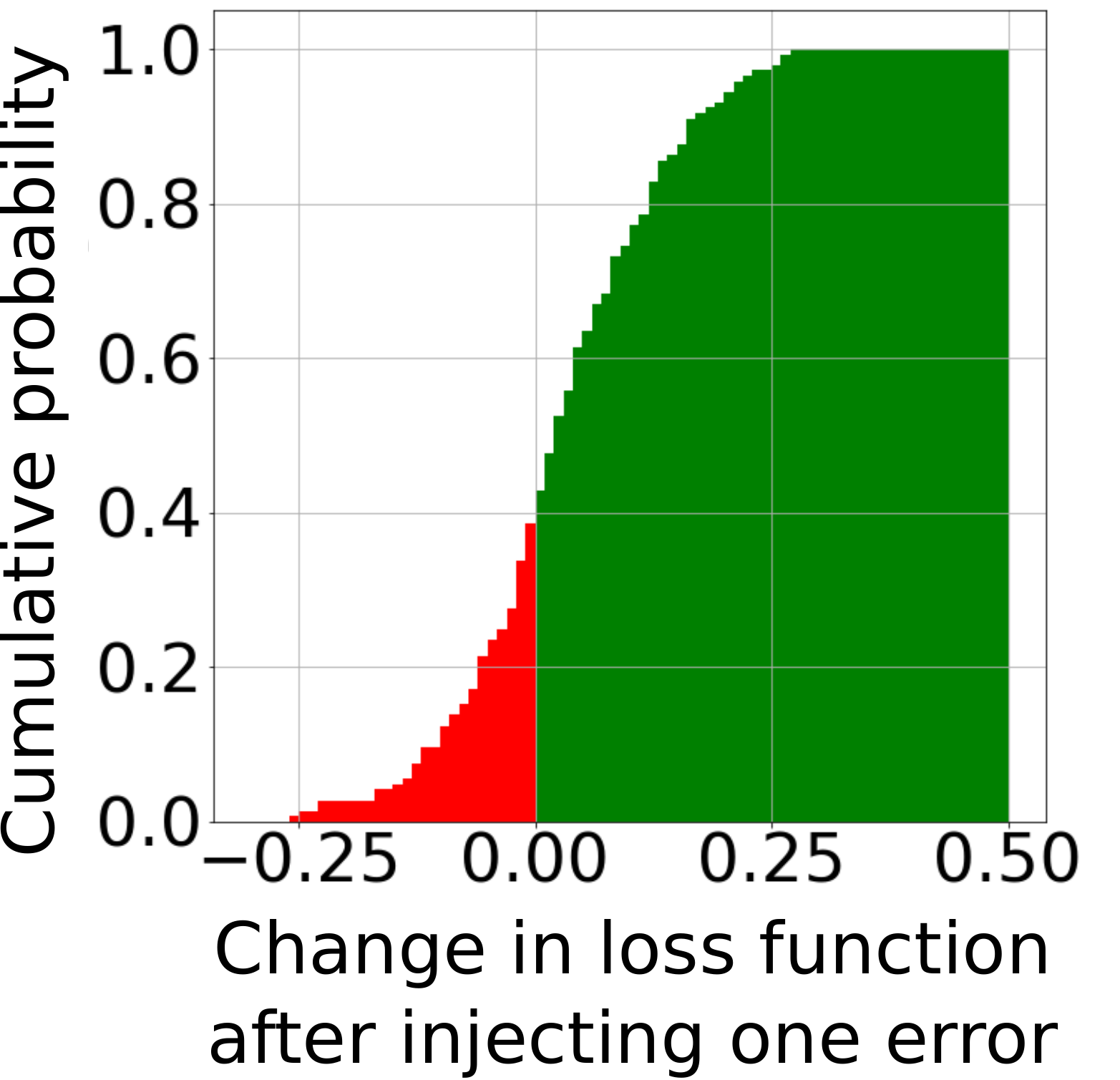} &
		\includegraphics[width=0.18\textwidth]{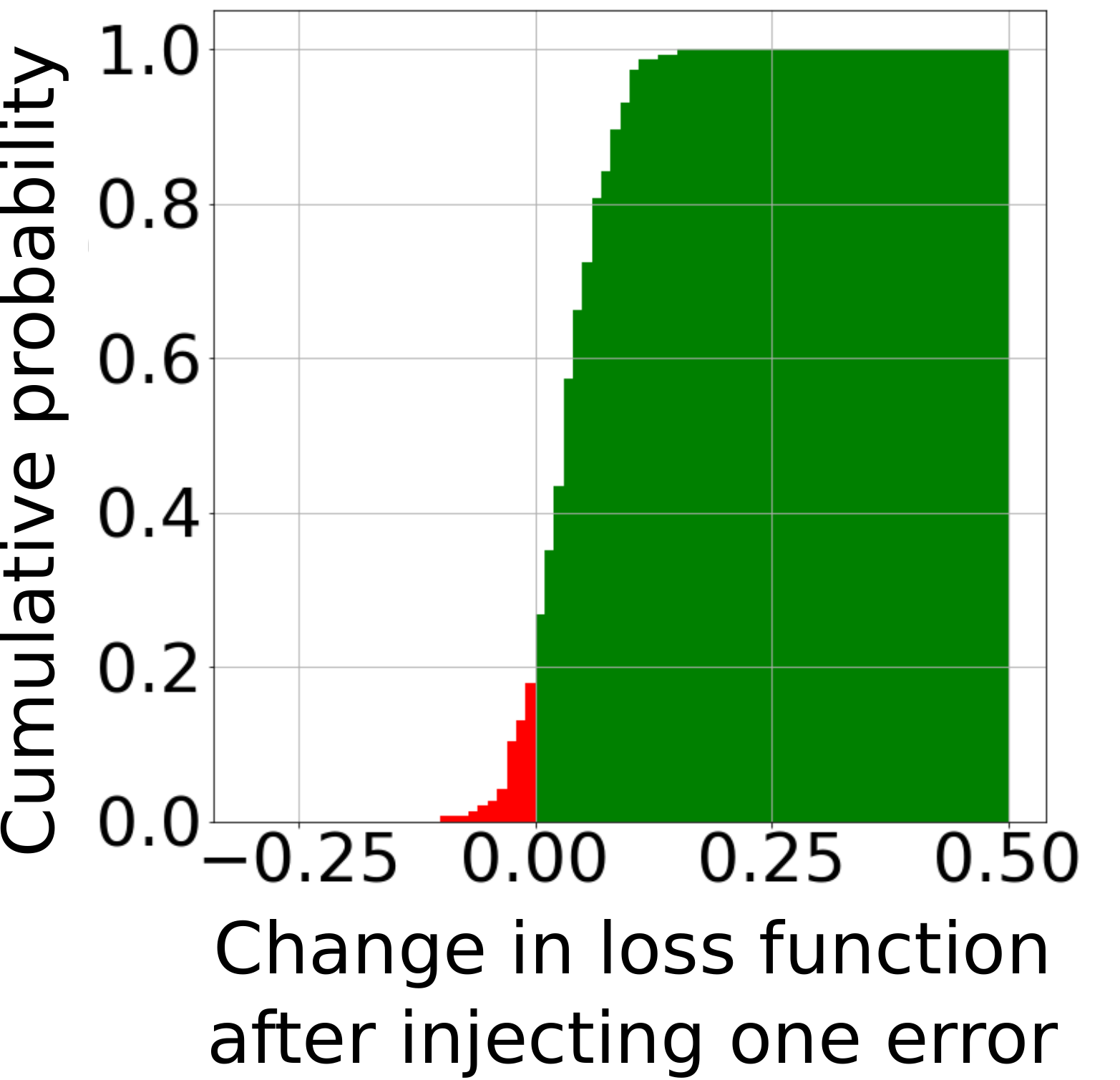} 
		\\
		(a) GT distance maps &
		(b) after error injection &
		(c) filtration by thresholding dist.\ maps&
		(d) our approach
	\end{tabular}
\vspace{-3mm}
\caption{\small {\bf Sensitivity of the topological loss term $C$ to the number of injected errors} (a) Ground truth distance maps of road networks. (b) Distance maps corrupted by introducing false roads and interruptions. We randomly injected one error at a time, obtaining corrupt distance maps with $30$ errors. We repeated this simulation $10$ times. (c,d) The cumulative distribution function of change in the loss term in response to injecting one error. In (c), $C$ is evaluated using the filtration by thresholding distance maps, whereas in (d) we use our filtration. The probability of decreasing the existing loss term by injecting additional errors is around $0.4$, whereas for our loss term it drops to $0.2$. We conclude that our loss term is more monotonic with respect to the error number.}
\label{fig:simulation}
\end{figure*}

%\vspace{-2mm}
\subsection{Correlation to the Number of Topological Errors}
%\vspace{-2mm}

\change{
We motivated our filtration technique by the fact that it introduces partial localization of {topological features} into the persistence diagrams and better spreads the diagrams across the plane. Here, we validate this on synthetic data to show that it correlates better with the number of errors injected into a distance map than the baseline loss, which is based solely on thresholding distance maps. To this end, we took two crops of ground truth road graphs of the \RTD{} dataset~\cite{Bastani18} and generated faulty synthetic distance maps by injecting thirty errors one at a time.  They were selected randomly and with equal probability between a road disconnection and a false interconnection. After each error injection, we evaluated the topological loss term $C$ of \eqref{eq:totloss} using either filtration by thresholding distance maps or our combined filtration. Ideally, we would expect $C$ to increase every time an error is added. Hence, we repeated the experiment ten times. For each crop, we plot the distribution of the increment in $C$ resulting from adding one error, when using the baseline loss in Fig.~\ref{fig:simulation}(c) and ours in Fig.~\ref{fig:simulation}(d). The parts of the distributions shown in green correspond to positive increments, which are what we expect, and those in red denote the negative ones, which are essentially erroneous. Note that the red parts are far smaller when using our loss than the baseline one.
}

%\vspace{-2mm}
\subsection{Performance in Training Deep Networks}
%\vspace{-2mm}
\change{
Having shown that our loss function captures topological correctness better than existing PH-based methods, we now compare the performance of deep nets trained with our and existing losses.}

%\vspace{-2mm}
\subsubsection{Datasets}
%\vspace{-2mm}

We experimented on three datasets.
\begin{itemize}[topsep=0pt]

\item \RTD{}. A dataset of high-resolution satellite images covering urban areas of forty cities in six countries~\cite{Bastani18}. The ground truth was obtained from OpenStreetMap. Like~\cite{Bastani18,Li18h,Yang19,Mosinska20}, we used  twenty five cities as the training set and the remaining fifteen as the test set. 

\item \MAS{}. The Massachusetts dataset~\cite{Mnih13} features both urban and rural neighborhoods, with many different kinds of roads ranging from small paths to highways. For a fair comparison to~\cite{Hu19b}, we split the data into three equal folds and performed a three-way cross validation.

\item \NEU{}. The dataset is a part of a proprietary 3D, 2-photon microscopy scan of a whole mouse brain. It contains 14 stacks of size $250 \times 250 \times 200$ voxels and a spatial resolution of $1.0 \times 0.3 \times 0.3$ ${\mu m}$. We used ten stacks for training and the remaining four for testing. 

\item \BRN{}. The dataset contains two 3D images of neurons in a mouse brain. The axons and dendrites have been outlined manually while viewing the sample under a microscope and the image has been captured later. The sample deformed in the meantime, resulting in a misalignment between the annotation and the image.
\changeSecond{To ensure that the test and training data comes from the same distribution, we split the two scans into stacks of $150\times200\times200$ voxels and a spatial resolution of $1~\mu m$, and randomly divided the resulting data set into a training set of twelve stacks and a test set of ten scans. }

\end{itemize}

%\vspace{-2mm}
\subsubsection{Methods tested}
%\vspace{-2mm}

To test the impact of our proposed filtration functions, we used the standard U-Net architecture~\cite{Ronneberger15}, with four blocks, each with two sequences of convolution-ReLU-batch normalization. Max-pooling in $2\times2$ windows followed each of the blocks. The initial feature size was set to $32$ and grew to $512$ in the smallest feature map in the network. We augmented the training data with vertical and horizontal flips and random rotations, and used the ADAM algorithm~\cite{Kingma15} with the learning rate set to $1e-4$. We then used different version of the $L_{tot}$ of Eq.~\ref{eq:totloss} we minimized to train the network. We tested the following as baselines:
\begin{itemize}[topsep=0pt]

 \item \CE{}. $L$ is the Cross Entropy loss for pixel classification and there is no topological discrepancy loss, that is, $\alpha=0.0$. \change{Binary masks are used as ground truth.}
 
 \item \MSE{}. $L$ is the mean squared error of the truncated distance to the closest foreground pixel, with no topological discrepancy loss.

 \item \HomoT{}. $L$ is the cross Entropy loss and we compute $C$ by thresholding pixel classification maps., as in{~\cite{Hu19b,Clough19,Clough20}}. 

 \item \HomoR{}. $L$ is the mean squared error and we compute $C$ by thresholding the truncated distance maps, as in~\cite{Wang20e}.
 
 \item \Ours{}.  $L$ is the mean squared error and we compute $C$ using our proposed filtration function. 

\end{itemize}
\change{Based on the results of the ablation study on the \MAS{} data set, presented in the supplementary material, in all our experiments with \HomoT{}, \HomoR{}, and \Ours{}, we set $\alpha=0.01$ and compute the loss in windows of size $64\times 64$ pixels.}
Like~\cite{Hu19b}, we limit the method to homolog{y classes} order $1$, that is, loops. This has two advantages. First, by convention, loops are created by the borders of the window, making disconnections in dead-ending roads or neurites detected as broken loops. Second, detection of homolog{y classes} is computationally expensive, and the time grows cubically with the number of pixels. In our current setup, computing the loss for a single window takes 0.5 seconds.
Similarly to~\cite{Hu19b}, we did not observe any performance gain due to using homolog{y classes} of order $0$---connected components---in addition to loops.

For completeness, we also compared our approach to recent techniques {\it not} relying on persistent homology: \Segm{} \cite{Bastani18}, \RTracer{} \cite{Bastani18}, \SegPath{} \cite{Mosinska20}, \RCNN{} \cite{Yang19},  \DRoad{} \cite{Mattyus17},  \PolyM{} \cite{Li18h},  \DMT{}~\cite{Hu21b}, and  \Malis{}~\cite{Oner21a}.
\Segm{}, \RTracer{}, \RCNN{}, and \PolyM{} do not explicitly enforce topology constraints, while the others do and are discussed in the related work section. 
{The outputs of these methods were shared by the authors directly with us or on the Internet, and we computed all the performance metrics.}

%\MK{Hyper-parameters of the filtration - range of slopes, window size, etc.}
%1) The loss is computed in windows sized $64\times 64$ pixels.
%2) Only homology order?rank? 1 (loops) are used. Because ...
%3) the coefficient $\alpha$ in~\eqref{eq:totloss} is set to $0.01$.
%\PF{The fact that the loss is computed over windows is not discussed in the method section. It should.}

%\vspace{-2mm}
\subsubsection{Performance metrics}
%\vspace{-2mm}

Comparing connectivity of segmentation masks is difficult, because the reconstructions rarely overlap with the ground truth, and often deviate from it significantly. There seems to be no consensus concerning the best evaluation technique; we found five connectivity-oriented metrics in concurrently published recent work. To provide an exhaustive evaluation, we used all of them.
\begin{itemize}[topsep=0pt]
 \item \APLS{}. Average Path Length Similarity aggregates relative length differences of shortest paths between pairs of corresponding points in the ground truth and predicted maps~\cite{VanEtten18}.
 
 \item \TLTS{} is a statistics of lengths of shortest paths between corresponding pairs of end points randomly selected in the predicted and ground-truth networks~\cite{Wegner13}. We report the fraction of paths with relative length difference within 5\%. 
 
 \item \Junc{}. It is a junction score that considers the number of roads intersecting at each junction~\cite{Bastani18}. It consists of road recall, averaged over the intersections of the ground-truth and road precision, averaged over the intersections of the prediction. We report the corresponding F1 score.
 
 \item \Betti{}. The Betti error~\cite{Hu19b} is an average absolute difference between the number of topological structures seen in the ground truth and predicted delineations. We take random patches sized $64\times 64$ from predictions, compute the number of 1-homology classes (loops) and compare the numbers computed for the prediction and the ground truth. We average this difference over $10$ trials. In practice, to compute the error we use the code made publicly available by the authors.
%a function of the number of topological structures seen in the ground-truth and prediction persistence diagrams. For homology, the $ k$-Betti number is the number of $k$- homology classes, where $k$ is the dimension: $0$ for connected components, $1$ for loops, $2$ for closed surfaces, and so on. In persistent homology, computing the number of homology classes and each step of the filtration yields Betti curves. A Betti curve consists of a function $f$ of the real line where $f(t)$ is the number of $k$-homology classes at scale $t$ of the filtration. The Betti error measures the difference between the Betti curves of the ground truth and the predicted persistence diagrams}. 
 
 \item \CCQ{} We complement the connectivity-oriented metrics with the most popular metric that measures spatial co-occurrence of annotated and predicted road pixels. The Correctness, Completeness and Quality are equivalent to precision, recall and intersection-over-union, with the definition of a true positive relaxed from spatial coincidence of prediction and annotation to co-occurrence within a distance of 5 pixels~\cite{Wiedemann98}. We report the Quality as our single-number metric.
 
\end{itemize}

% !TEX root = ../top.tex
% !TEX spellcheck = en-US

\begin{table*}[htb!]
\small
\centering
\caption{
\change{Validation results on the \MAS{} dataset.}
Our loss function outperforms all PH-based loss functions.
We report means and standard deviations over three independent training runs.
\label{tab:results-mas}
}
%	\vspace{-4mm}
	\begin{tabular}{@{} p {0.28\columnwidth} @{}>{\centering\arraybackslash}p{1.6cm}>{\centering\arraybackslash}p{1.6cm}>{\centering\arraybackslash}p{1.6cm} >{\centering\arraybackslash}p{1.6cm}@{}>{\centering\arraybackslash}p{0.2cm}@{}>{\centering\arraybackslash}p{2.2cm} @{} }
		\cmidrule{2-7}
		
		& \multicolumn{4}{c}{Connectivity-oriented} && pixel-based \\
		
		\cmidrule{2-5}
		\cmidrule{7-7}
		
		Method &    \APLS{} $\uparrow$ &       \TLTS $\uparrow$ &      \Junc $\uparrow$ &         \Betti $\downarrow$ &&        \CCQ $\uparrow$ \\
		\cmidrule{1-7}
		
		\CE{} &
		$60.9 \pm 3.9$  &   $41.6 \pm 4.1$    &    $72.0 \pm 2.7$  &  $3.12 \pm 0.6$  &&    $66.9 \pm 2.6$ \\
		\MSE{} &
		$61.3 \pm 3.7$ &    $41.9 \pm 4.2$    &    $71.9 \pm 2.9$  &  $3.09 \pm 0.7$ &&    $67.3 \pm 2.3$ \\
		
		\cmidrule{1-7}
		
		\DMT{} &
		$64.7 \pm 2.9$ &    $45.8 \pm 2.8$    &    $80.6 \pm 2.4$  &  $\textbf{0.99} \pm 0.4$  &&    $74.9 \pm 1.9$ \\
		\Malis{}&
		$\textbf{73.4} \pm 3.6$ &   $ \textbf{53.2} \pm 4.4$    &    $\textbf{81.4} \pm 1.9$  &  $1.29 \pm 0.5$ &&    $\textbf{75.8} \pm 2.2$ \\
		
		\cmidrule{1-7}
		
		\HomoT{}&
		$62.5 \pm 1.9$ &    $42.1 \pm 1.9$    &    $74.2 \pm 1.7$  &  $1.28 \pm 0.3$  &&    $69.3 \pm 1.9$ \\
		\HomoR{} &
		$65.0 \pm 2.2$ &   $45.6 \pm 1.8$     &  $76.9 \pm 1.9$   &  $1.09 \pm 0.2$  &&    $71.8 \pm 2.1$ \\
		\Ours{}  & 
		$\textbf{68.7} \pm 1.2$ &    $\textbf{50.6} \pm 2.3$    &    $\textbf{79.2} \pm 2.6$  &  $\textbf{0.90} \pm 0.3$  &&    $\textbf{74.9} \pm 1.8$ \\
		
		\cmidrule{1-7}
		
	\end{tabular}
\end{table*}

% !TEX root = ../top.tex
% !TEX spellcheck = en-US

\begin{table*}[t]
\small
	\centering
	\caption{
		Our loss function outperforms all PH-based loss functions on the \RTD{} dataset.
		We report means and standard deviations over cities from the test set. 
		\label{tab:results-rt}
	}
	\vspace{-4mm}
	\begin{tabular}{@{} p{0.28\columnwidth} @{}>{\centering\arraybackslash}p{1.6cm}>{\centering\arraybackslash}p{1.6cm}>{\centering\arraybackslash}p{1.6cm} >{\centering\arraybackslash}p{1.6cm}@{}>{\centering\arraybackslash}p{0.2 cm }@{}>{\centering\arraybackslash}p{2.2cm} @{} }
		\cmidrule{2-7}
		
		& \multicolumn{4}{c}{Connectivity-oriented} && pixel-based \\
		
		\cmidrule{2-5}
		\cmidrule{7-7}
		
		Method &    \APLS{} $\uparrow$ &       \TLTS $\uparrow$ &      \Junc $\uparrow$ &         \Betti $\downarrow$ &&        \CCQ $\uparrow$ \\
		\cmidrule{1-7}
		
		\CE{} &
		$63.4 \pm 1.6$ &  $37.5 \pm 1.9$    &  $78.0 \pm 1.0$  &  $3.08 \pm 0.6$  &&  $59.7 \pm 2.2$ \\
		\MSE{} &
		$66.3 \pm 1.9$ &  $40.0 \pm 2.0$    &  $77.5 \pm 1.3$  &  $2.99 \pm 0.5$   && $59.5 \pm 1.9$ \\

		\cmidrule{1-7}
		
		\Segm{} &
		$62.5 \pm 1.5$ &   $33.0 \pm 1.6$ &  $78.2 \pm 1.5$ & $3.04 \pm 0.6$ &&  $54.4 \pm 1.0$ \\ 
		\RTracer{}&
		$59.1 \pm 0.8$ &    $40.6 \pm 1.5$ &  $81.2 \pm 1.6$ & $2.85 \pm 0.7$ && $47.8 \pm 1.6$ \\
		\SegPath{} &
		$68.1 \pm 1.4$ &    $46.5 \pm 1.7$ &  $75.4 \pm 1.3$ & $2.31 \pm 0.4$ && $54.0 \pm 1.4$ \\
		\RCNN{} &
		$48.2 \pm 1.6$ &    $18.4 \pm 1.9$ & $75.9 \pm 1.4$&  $3.25 \pm 0.7$ &&    $62.8 \pm 1.5$ \\
		\DRoad{} &
		$24.6 \pm 2.2$ &   $6.4 \pm 0.9$ &   $51.4 \pm 1.5$ &  $4.95 \pm 1.0$  &&   $43.6 \pm 2.0$ \\
		\PolyM{} &
		$61.3 \pm 2.3$ &   $31.5 \pm 1.9$ &   $80.0 \pm 1.2$ & $2.90 \pm 0.4$ &&   $35.7 \pm 1.4$ \\
		\Malis{}  &
		$\textbf{75.4} \pm 1.6$ &    $\textbf{49.6} \pm 1.4$    &    $\textbf{82.6} \pm 0.6$  &  $\textbf{1.30} \pm 0.4$  &&    $\textbf{68.4}\pm 0.9$ \\
				
		\cmidrule{1-7}
		
		\HomoT{} &
		$67.3 \pm 1.7$ &    $42.3 \pm 1.1$    &   $78.7 \pm 0.9$   &  $1.32 \pm 0.3$  &&    $61.9 \pm 1.9$ \\
		\HomoR{} &
		$69.9 \pm 1.6$  &    $45.1 \pm 1.4$   &    $79.6 \pm 1.3$  &  $1.07 \pm 0.3$  &&    $63.2 \pm 1.6$ \\
		\Ours{}  & 
		$\textbf{73.8} \pm 1.8$ &    $\textbf{47.8} \pm 0.9$    &    $\textbf{81.3} \pm 1.6$  &  $\textbf{0.89} \pm 0.2$  &&    $\textbf{66.3} \pm 1.7$ \\
		
		\cmidrule{1-7}
		
	\end{tabular}
\end{table*}

% !TEX root = ../top.tex
% !TEX spellcheck = en-US

\begin{table*}[!h]
\small
	\centering
	\caption{
		Comparative results on the \NEU{} dataset.
		Our loss outperforms all the baselines.
		We report means and standard deviations over three independent training runs.
		\label{tab:results-neurons}
	}
	\vspace{-4mm}
	\begin{tabular}{@{} p {0.28\columnwidth} @{}>{\centering\arraybackslash}p{1.6cm}>{\centering\arraybackslash}p{1.6cm}>{\centering\arraybackslash}p{1.6cm}@{}>{\centering\arraybackslash}p{0.2cm}@{}>{\centering\arraybackslash}p{2.2cm} @{} }
		\cmidrule{2-6}
		
		& \multicolumn{3}{c}{Connectivity-oriented} && pixel-based \\
		
		\cmidrule{2-4}
		\cmidrule{6-6}
		
		Method &    \APLS{} $\uparrow$ &       \TLTS $\uparrow$ &    \Betti $\downarrow$ &&        \CCQ $\uparrow$ \\
		\cmidrule{1-6}
		\CE{} &
		$79.9 \pm 1.5$ &    $80.8 \pm 2.2$    &  $2.33 \pm 0.6$  &&    $90.6 \pm 2.0$ \\
		\MSE{} &
		$80.2 \pm 1.6$ &   $80.9 \pm 2.0$    &  $2.31 \pm 0.7$  &&    $90.4 \pm 1.9$ \\
		
		\cmidrule{1-6}
		
		\HomoT{}&
		$83.5 \pm 1.0$ &   $82.1 \pm 1.7$    &  $1.06 \pm 0.2$  &&    $91.2 \pm 1.8$\\
		\HomoR{} &
		$85.4 \pm 1.2$ &   $83.4 \pm 1.5$    &  $0.91 \pm 0.2$  &&    $92.5 \pm 1.6$ \\
		\Ours{}  & 
		$\textbf{86.9} \pm 1.1$ &    $\textbf{85.2} \pm 1.9$    &  $\textbf{0.80} \pm 0.2$  &&    $\textbf{93.3} \pm 1.9$ \\
		
		\cmidrule{1-6}
		
	\end{tabular}
\end{table*}

% !TEX root = ../top.tex
% !TEX spellcheck = en-US

\begin{table*}[!h]
\small
	\centering
	\caption{
		{Comparative results on the \BRN{} dataset.
		Our loss outperforms all PH-based losses.
		Means and standard deviations over three independent training runs as presented.
		\label{tab:results-brain}}
	}
	\vspace{-4mm}
	\begin{tabular}{@{} p {0.28\columnwidth} @{}>{\centering\arraybackslash}p{1.6cm}>{\centering\arraybackslash}p{1.6cm}>{\centering\arraybackslash}p{1.6cm}@{}>{\centering\arraybackslash}p{0.2cm}@{}>{\centering\arraybackslash}p{2.2cm} @{} }
		\cmidrule{2-6}
		
		& \multicolumn{3}{c}{Connectivity-oriented} && pixel-based \\
		
		\cmidrule{2-4}
		\cmidrule{6-6}
		
		Method &    \APLS{} $\uparrow$ &       \TLTS $\uparrow$ &    \Betti $\downarrow$ &&        \CCQ $\uparrow$ \\
		\cmidrule{1-6}
		\CE{} &
		$65.8 \pm 1.8$ &    $63.6 \pm 1.3$    &  $2.89 \pm 0.4$  &&    $70.4 \pm 1.9$ \\
		\MSE{} &
		$66.0 \pm 1.6$ &    $63.9 \pm 1.4$    &  $2.92 \pm 0.5$  &&    $70.6 \pm 1.8$ \\
		
		\cmidrule{1-6}
		
		\HomoT{}&
		$67.6 \pm 1.5$ &    $65.3 \pm 1.0$    &  $1.39 \pm 0.2$  &&    $71.5 \pm 1.4$\\
		\HomoR{} &
		$70.5 \pm 1.5$ &    $68.8 \pm 0.9$    &  $1.22 \pm 0.3$  &&    $72.6 \pm 1.3$ \\
		\Ours{}  & 
		$\textbf{73.4} \pm 1.4$ &    $\textbf{70.1} \pm 1.1$    &  $\textbf{1.06} \pm 0.2$  &&    $\textbf{73.2} \pm 1.2$ \\
		
		\cmidrule{1-6}
		
	\end{tabular}
\end{table*}

%\vspace{-2mm}
\subsubsection{Comparative Results}
%\vspace{-2mm}

\change{We present validation results for the \MAS{} data set in Tab.~\ref{tab:results-mas}, and test results for the \RTD{} data set in Tab.~\ref{tab:results-rt}.}
Our method outperforms the other methods based on Persistent Homology, which demonstrates that our approach to filtering is truly effective. 
It also outperforms the other 2D tracing algorithms targeted at handling aerial images,  \RTracer{}, \SegPath{}, \DRoad{}, and \PolyM{}, with the exception of \Malis{} that does marginally better. This is presumably because  \Malis{} explicitly penalizes each disconnection of the prediction, whereas a persistence diagram is a lossy topological descriptor that may fail to penalize some errors. However, \Malis{} does not naturally extend to 3D data, whereas our method does.  On the 3D \NEU{} data set, it outperforms the competing algorithms, as evidenced by the \change{test} results shown in Tabs~\ref{tab:results-neurons} and~\ref{tab:results-brain}. We provide qualitative results in the supplementary material. 

% Notably, the \Malis{}, \RTracer{}, \SegPath{}, \DRoad{}, and \PolyM{} are targeted at road segmentation in 2D aerial images, and their adaptation to 3D is not trivial. 

% !TEX root = ../top.tex
% !TEX spellcheck = en-US

%\vspace{-2mm}
\section{Conclusion}
%\vspace{-2mm}

We demonstrated a fault in the design of existing methods to employ Persistent Homology to train deep networks in delineating curvilinear structutres: by using inadequate filtration functions, they severely reduce the information content of the persistence diagrams, hampering performance of the trained network. We proposed an improved approach, based on combining filtration by thresholding with the height function, that increases the descriptive power of the diagrams, and gives PH a place among the best-performing methods to train topologically accurate deep networks.

\change{
The proposed approach is limited by the need to randomly select the parameters of the height function at each training iteration, because
some orientations of the height axis might result in a failure to detect topological errors, or provoke erroneous matches between the persistence diagrams of the prediction and the ground truth.
We therefore plan to investigate the use of multipersistence~\cite{Carlsson09} for less random and more effective supervision.
}
\change{
Another limitation stems from the fact that our loss function has sparse gradients that only depend on values at pixels that are critical for emergence and disappearance of {topological features}. This limits robustness and our future work will focus on developing topological descriptors with more smooth gradients.
}
\change{
While our loss function improves the topological correctness of the segmentation masks, some bio-medical applications require full confidence of correctness of anatomy models, which current methods cannot guarantee. This also motivates us to investigate the use of topological methods to highlight the regions of the segmentation masks that require manual correction, thereby facilitating proof-reading of segmentation results. 
}

\section*{Acknowledgement}

This work was supported in part by the Swiss National Science Foundation Sinergia grant no.\ 177237
and by the FWF Austrian Science Fund Lise Meitner grant no.\ M3374.
\vspace{-1em}

\bibliographystyle{IEEEtran}
\bibliography{string,vision,learning,biomed,topo,misc}
\end{document}